\documentclass{article}

\usepackage{arxiv}
\usepackage[utf8]{inputenc} 
\usepackage[T1]{fontenc}    
\usepackage{hyperref}       
\usepackage{url}            
\usepackage{booktabs}       
\usepackage{amsfonts}       
\usepackage{nicefrac}       
\usepackage{microtype}      
\usepackage{lipsum}		
\usepackage{graphicx}
\usepackage{natbib}
\usepackage{doi}

\title{Cyclical Weight Consolidation: Towards Solving Catastrophic Forgetting in Serial Federated Learning}

\author{{\hspace{1mm}Haoyue Song} \\
	National Institute for Data Science in Health and Medicine\\
	Xiamen University\\
	\texttt{hysong@stu.xmu.edu.cn} \\
	\And
	\hspace{1mm}Jiacheng Wang \\
	Department of Computer Science\\
	Xiamen University\\
	\texttt{jiachengw@stu.xmu.edu.cn} \\
	\And
	\hspace{1mm}Liansheng Wang \\
	Department of Computer Science\\
	Xiamen University\\
	\texttt{lswang@xmu.edu.cn} \\
}

\hypersetup{
pdftitle={A template for the arxiv style},
pdfsubject={q-bio.NC, q-bio.QM},
pdfauthor={David S.~Hippocampus, Elias D.~Striatum},
pdfkeywords={First keyword, Second keyword, More},
}

\begin{document}
\maketitle

\renewcommand{\shorttitle}{Cyclical Weight Consolidation}

\begin{abstract}
Federated Learning (FL) has gained attention for addressing data scarcity and privacy concerns. While parallel FL algorithms like FedAvg exhibit remarkable performance, they face challenges in scenarios with diverse network speeds and concerns about centralized control, especially in multi-institutional collaborations like the medical domain. 
Serial FL presents an alternative solution, circumventing these challenges by transferring model updates serially between devices in a cyclical manner. Nevertheless, it is deemed inferior to parallel FL in that (1) its performance shows undesirable fluctuations, and (2) it converges to a lower plateau, particularly when dealing with non-IID data. The observed phenomenon is attributed to catastrophic forgetting due to knowledge loss from previous sites. In this paper, to overcome fluctuation and low efficiency in the iterative learning and forgetting process, we introduce cyclical weight consolidation (CWC), a straightforward yet potent approach specifically tailored for serial FL. 
CWC employs a consolidation matrix to regulate local optimization. This matrix tracks the significance of each parameter on the overall federation throughout the entire training trajectory, preventing abrupt changes in significant weights. During revisitation, to maintain adaptability, old memory undergoes decay to incorporate new information. Our comprehensive evaluations demonstrate that in various non-IID settings, CWC mitigates the fluctuation behavior of the original serial FL approach and enhances the converged performance consistently and significantly. The improved performance is either comparable to or better than the parallel vanilla.
\end{abstract}

\section{Introduction}

\begin{figure}[t]
\centering
    \includegraphics[width=0.7\linewidth]{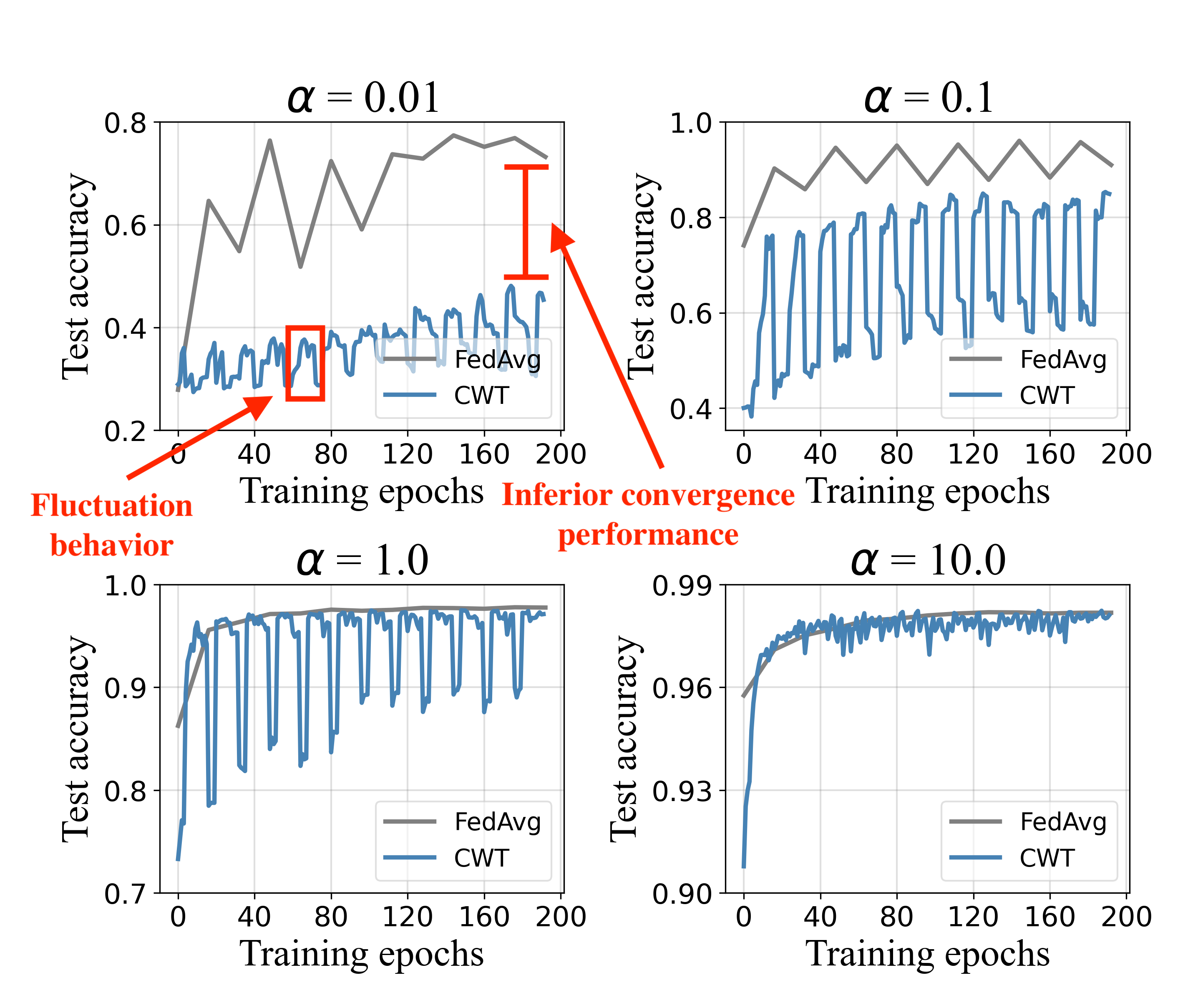}
    \caption{
    A comparison between CWT and FedAvg on MNIST reveals the inferiority of CWT in causing undesirable fluctuations in performance and converging to a lower plateau when confronted with non-IID data. We present figures under various Dirichlet concentration parameters $\alpha$ (i.e., 0.01, 0.1, 1.0, and 10.0, arranged from top to bottom and left to right). A smaller $\alpha$ corresponds to greater data heterogeneity.
    }
    \label{fig:motivation}
    \vspace{-2mm}
\end{figure}

In recent years, federated learning (FL)~\cite{fl1,fl2,wang2022personalizing,wang2023feddp} has shown remarkable performance across various domains, including medicine, finance, and IoT. FL addresses the challenge of data scarcity among isolated clients, often bound by privacy regulations, by training a global model in a distributed manner, enhancing model performance while respecting data ownership. These achievements are largely credited to parallel FL algorithms like FedAvg~\cite{fedavg}, which involve a central server coordinating the aggregation of updated weights from participating clients iteratively until convergence. However, it's vital to acknowledge that parallel FL may not be the optimal solution for all scenarios, especially when logistical challenges~\cite{cwt} and concerns regarding a centralized third party~\cite{proxyFL} are paramount. The former scenario arises when collaborators have significantly different network connection speeds or deep learning hardware, while the latter situation is particularly undesirable in multi-institutional collaborations. This is especially true in the medical domain, where each hospital may insist on autonomy over its own model to comply with regulations.

To circumvent these challenges, serial FL (i.e. CWT~\cite{cwt}) presents as an alternative solution by transferring model updates serially between devices in a cyclical manner. Nevertheless, it is deemed inferior to parallel FL in that (1) its performance shows undesirable fluctuations, and (2) it converges to a lower plateau, particularly when dealing with non-IID data~\cite{sheller1,sheller2}. This phenomenon (Fig.~\ref{fig:motivation}) is attributed to \textbf{catastrophic forgetting}~\cite{cf1,cf2,cf3}, where previously acquired knowledge from prior sites gradually fades as the model assimilates new information. Consequently, when the learning process at a new site concludes, the model performs well in the new data environment but at the cost of deteriorating performance on datasets from previously visited sites. The repetitive cycle of learning and forgetting leads to oscillating learning dynamics, significantly diminishing efficiency and resulting in undesirable convergence performance.

To address catastrophic forgetting induced by heterogeneous data distribution in serial FL, prior research has employed simple heuristics such as balanced mini-batch sampling or a weighted cross-entropy loss~\cite{cwt2}. To the best of our knowledge, no existing work has delved into the intricate mechanism of catastrophic forgetting within the cyclical weight transfer process and provided a targeted solution. Furthermore, although serial FL bears resemblance to continual learning (CL)~\cite{cl1,cl2,cl3}, where catastrophic forgetting is also a primary concern, techniques from CL cannot be directly applied to serial FL. This disparity stems from the inherent differences between the two tasks: while CL aims to sequentially learn and adapt to new tasks over time while preserving knowledge from past experiences, serial FL distinguishes itself by cyclically revisiting previously accessed sites. This unique characteristic necessitates the development of a new anti-catastrophic forgetting design that can fully leverage this cyclic revisitation feature, thereby enhancing overall performance.

In this paper, we present a straightforward yet potent approach, termed Cyclical Weight Consolidation (CWC), to address the challenge of catastrophic forgetting in serial federated learning. Instead of granting the model unrestricted autonomy to update itself, CWC opts for local optimization under the regularization of a consolidation matrix. This matrix encapsulates knowledge from previously visited sites, thereby preventing significant weights from undergoing abrupt changes. Following the optimization at the current site, the matrix updates itself by assimilating important weight estimations specific to that site. Given the cyclical revisitation of all sites' data, the consolidation matrix undergoes attenuation at the onset of each new communication round, preventing permanent adherence to out-of-date knowledge and preserving the model's adaptability. Our comprehensive evaluations illustrate that, across various non-IID settings, CWC alleviates the fluctuation behavior exhibited by the original serial FL approach and consistently and significantly enhances convergence performance. The improved performance is either comparable to or surpasses that of the parallel vanilla. In a nutshell, our contributions are summarized as follows:
\begin{itemize}
    \item We present our algorithm for serial federated learning (CWC). Through experimental demonstrations, we show that by utilizing the consolidation matrix to regularize the local optimization process, CWC significantly mitigates the issue of catastrophic forgetting stemming from non-IID data distributions.
    \item Extensive experiments highlight the superiority of CWC over serial federated learning vanilla (CWT). CWC effectively mitigates the fluctuation behavior observed in the original serial FL approach and consistently and significantly enhances convergence performance. Moreover, CWC has the capability to boost convergence by intensifying local computation, a feature not achievable in CWT. 
    \item Extensive experiments affirm that CWC either matches or surpasses the performance of FedAvg. To the best of our knowledge, this marks the first instance where the serial FL method achieves performance comparable to the parallel vanilla approach in non-IID settings.
\end{itemize}
\section{Related Work}

\subsection{Serial Federated Learning}

Federated learning has been proposed as a solution to address privacy concerns in distributed learning environments. In contrast to the pioneering federated method, FedAvg~\cite{fedavg}, which trains a global model by aggregating local model parameters concurrently, serial FL methods, exemplified by CWT~\cite{cwt} and Split learning~\cite{splitlearning}, train each local model sequentially rather than simultaneously. This sequential process continues iteratively, allowing the model to be refined over time as it incorporates insights from different clients in each round. However, serial FL is highly sensitive to data heterogeneity, leading to fluctuation behavior and lower convergence performance due to catastrophic forgetting~\cite{cf1,cf2,cf3}. Consequently, serial FL is considered an inferior paradigm compared to FedAvg~\cite{sheller1,sheller2}. To address these challenges, ~\cite{cwt2} employs proportional local training iterations, cyclical learning rates for sample size variability, and locally weighted mini-batch sampling, along with cyclically weighted loss for label distribution variability. While these simple heuristics demonstrate some effectiveness, they are only applicable in settings with moderate heterogeneity and offer limited improvement. They are inadequate for handling highly heterogeneous distributions such as partitions with mutual exclusive classes~\cite{SI,classimbalance1,classimbalance2}. To the best of our knowledge, no existing work has thoroughly explored the intricate mechanism of catastrophic forgetting within the cyclical weight transfer process and provided a targeted solution.

\subsection{Catastrophic Forgetting}

Catastrophic forgetting~\cite{cf1,cf2,cf3}, also known as catastrophic interference~\cite{ci1,ci2,ci3}, refers to a phenomenon in machine learning and artificial intelligence where a model that has been trained on one task significantly forgets or loses its ability to perform well on a previous task when it is trained on a new, unrelated task. In other words, as the model learns new information, it unintentionally erases or overwrites previously learned knowledge, leading to a decline in performance on earlier tasks. This issue is especially prominent in continual or lifelong learning scenarios~\cite{cl1,cl2,cl3,cl4,cl5}, where a model needs to adapt to new tasks over time. Proposed CL methods to overcome catastrophic forgetting can be broadly classified into three categories: rehearsal, architectural, and regularization methods. Rehearsal methods~\cite{rehearsal1,rehearsal2,rehearsal3,rehearsal4} involve storing and replaying essential samples from previous tasks during the training of a model on new tasks. This periodic revisitation of past experiences allows the model to retain knowledge about earlier tasks, thereby mitigating the risk of forgetting. However, rehearsal methods are not applicable in the context of serial FL, as they violate privacy requirements. Architectural methods~\cite{arch1,arch2,arch3} focus on designing neural networks with separate modules or compartments dedicated to different tasks. This modular approach ensures that knowledge acquired for each task is stored in distinct compartments, reducing interference between tasks and minimizing catastrophic forgetting. Regularization techniques~\cite{EWC,SI,reg3,reg4} introduce constraints into the learning process to prevent the model from becoming overly specialized on the current task and forgetting previously acquired information. Despite the effectiveness of these methods in CL tasks, their direct application to serial FL is not feasible. This discrepancy stems from the inherent differences between the two tasks: while CL focuses on sequential learning and adaptation to new tasks over time while preserving knowledge from past experiences, serial FL stands out by cyclically revisiting previously accessed sites. This distinctive characteristic calls for the development of a novel anti-catastrophic forgetting design that can fully exploit this cyclic revisitation feature, thereby enhancing overall performance.

\section{Approach}

\begin{figure*}[t]
\centering
    \includegraphics[width=\textwidth]{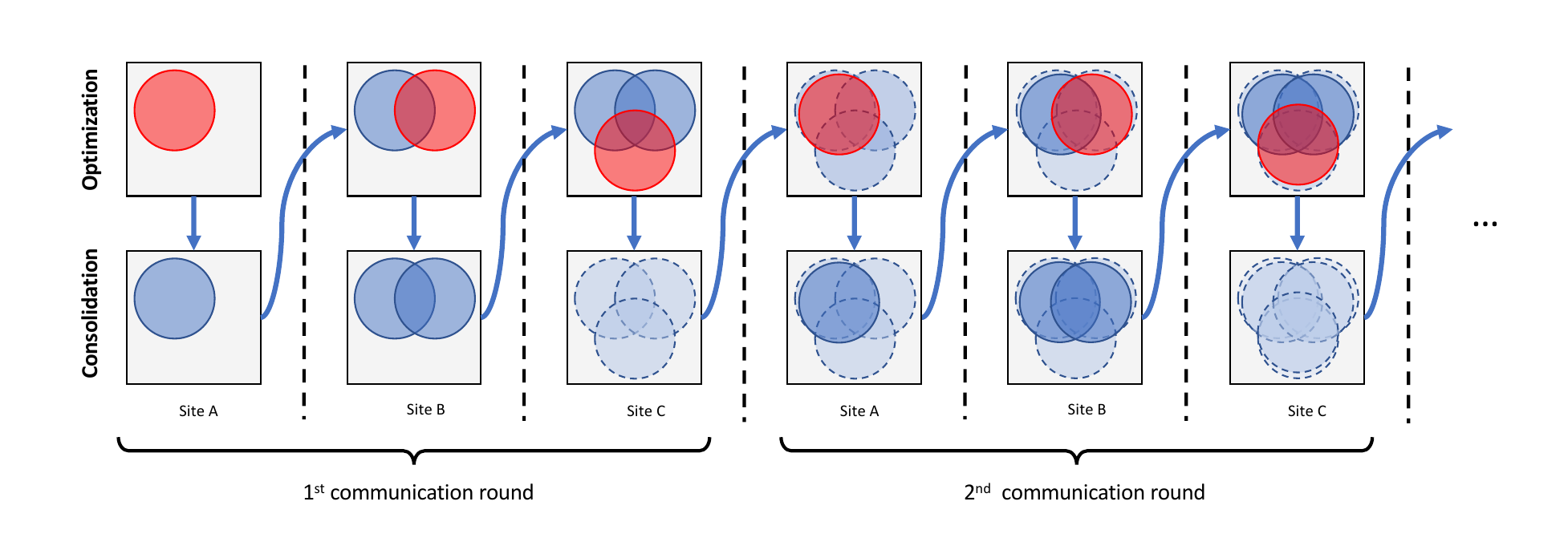}
    \caption{
   The procedure of our proposed cyclical weight consolidation (CWC). The red circle indicates the parameters being optimized during the optimization process, and the blue circle refers to the parameters being consolidated. The light blue circle with a dashed border indicates the attenuated consolidated parameters. And the rectangular box denotes the entire parameter set of the model.
   }
    \label{fig:cwc}
    \vspace{-2mm}
\end{figure*}

In this section, we begin by formally defining serial federated learning. Following this, we delve into the concept of cyclical weight consolidation, elucidating its underlying mechanisms that aid in overcoming catastrophic forgetting in serial FL.

\subsection{Problem Formulation}

In the context of federated learning involving a total of $K$ clients, the optimization objective is to learn a optimal global model $\theta$ which can generalize well to all the $K$ clients' datasets $\{D_k\}_{k=1}^K$: 
\begin{equation}
    min \space F(\theta) := \frac{1}{K} \sum_{k=1}^{K} \mathcal{L}_k \label{eq1}
\end{equation}
where $\theta \in \mathbb{R}^d$ encodes the parameters of the global model and $\mathcal{L}_k$ represents the loss incurred by client \(k\) when fitting the model $\theta$ to its local dataset \(D_k\):
\begin{equation}
    \mathcal{L}_k := E_{(x,y)\sim{D_k}}[\mathcal{L}_k(\theta; D_k)] \label{eq2}  
\end{equation}
In the absence of a centralized dataset, serial federated learning (i.e. CWT), approximates the objective through a sequential and cyclical training approach. During each training round, CWT sequentially trains the global model $\theta$ on individual local clients using their respective local data for a specified number of epochs. Subsequently, the global model is transferred to the next client for a similar training process. This cycle continues until all local clients have undergone training. The process iterates through the clients until convergence or the completion of a predetermined number of communication rounds.

The native CWT algorithm encounters challenges in serial federated learning scenarios characterized by heterogeneous client data distributions. Catastrophic forgetting manifests shortly after the model is transferred to the next site and begins adapting to the new data distribution. The previously acquired knowledge from the prior sites gradually fades as the model assimilates new information. Consequently, when the learning process at the new site concludes, the model excels in the new data environment but at the expense of deteriorating performance on datasets from previously visited sites. Hence, our goal is to reduce the negative effects of catastrophic forgetting.

\subsection{Cyclical Weight Consolidation}
We depict the process of cyclical weight consolidation in Fig.~\ref{fig:cwc}. The concept behind our proposed algorithm is that, instead of granting the model absolute freedom to update itself in the new data environment, we opt to conduct local optimization under the regularization of a consolidation matrix $C^{k,r}$, where $k$ and $r$ index the $k$-th client and $r$-th communication round, respectively. The consolidation matrix tracks the significance of each parameter on the overall federation (Eq.~(\ref{eq1})) throughout the entire training trajectory. Therefore, it resides in the same vector space as the model parameter $\theta$. With CWC, the optimization term (Eq.~(\ref{eq2})) is now extended to:
\begin{equation}
    \mathcal{L'}_k = \mathcal{L}_k + \sigma \sum_{i} C_i^{k,r} (\theta_i - \theta_i^{prev})^2 
\end{equation}
where $\sigma$ is the consolidation factor, balancing between old acquired knowledge and the new one, $\theta^{prev}$ represents the updated parameters transferred from the previous site and $i$ labels each parameter. 

The calculation of the consolidation matrix involves obtaining accurate important weight estimation for a specific task, an aspect extensively studied in the context of continual learning. We can leverage the Fisher information matrix (i.e., EWC~\cite{EWC}) or compute the sensitivity of the task loss to each parameter (i.e., SI~\cite{SI}) to find a good estimation. Through either method, we can obtain the important weight estimation $E_i^{k,r}$, which quantifies the significance of each parameter $i$ concerning site $k$ at round $r$. With the obtained important weight estimation $E_i^{k,r}$, we calculate $C_i^{k,r}$ as follows:
\begin{equation}
    C_i^{k,r} = \sum_{p=1}^{k-1} E_i^{p,r}, \; for \; 2 \leq k \leq K 
\end{equation}
Note that in this step, we aggregate all relevant important weight estimations from prior sites to compute the consolidation matrix for the current site. For the initial optimization, we set $C^{k=1,r=1}_i = 0$ as there is no previously consolidated information available. Since the algorithm cyclically revisits the $K$ sites, we attenuate the consolidation matrix at the start of a new communication round. This ensures that the model is not overly influenced by the out-of-date information, preserving its adaptability: 
\begin{equation}
    C_i^{1,r+1} = \gamma (C_i^{K,r} + E_i^{K,r})
\end{equation}
where $\gamma$ is the attenuation rate.

\section{Experiments}

\begin{figure*}[t]
\centering
    \includegraphics[width=\textwidth]{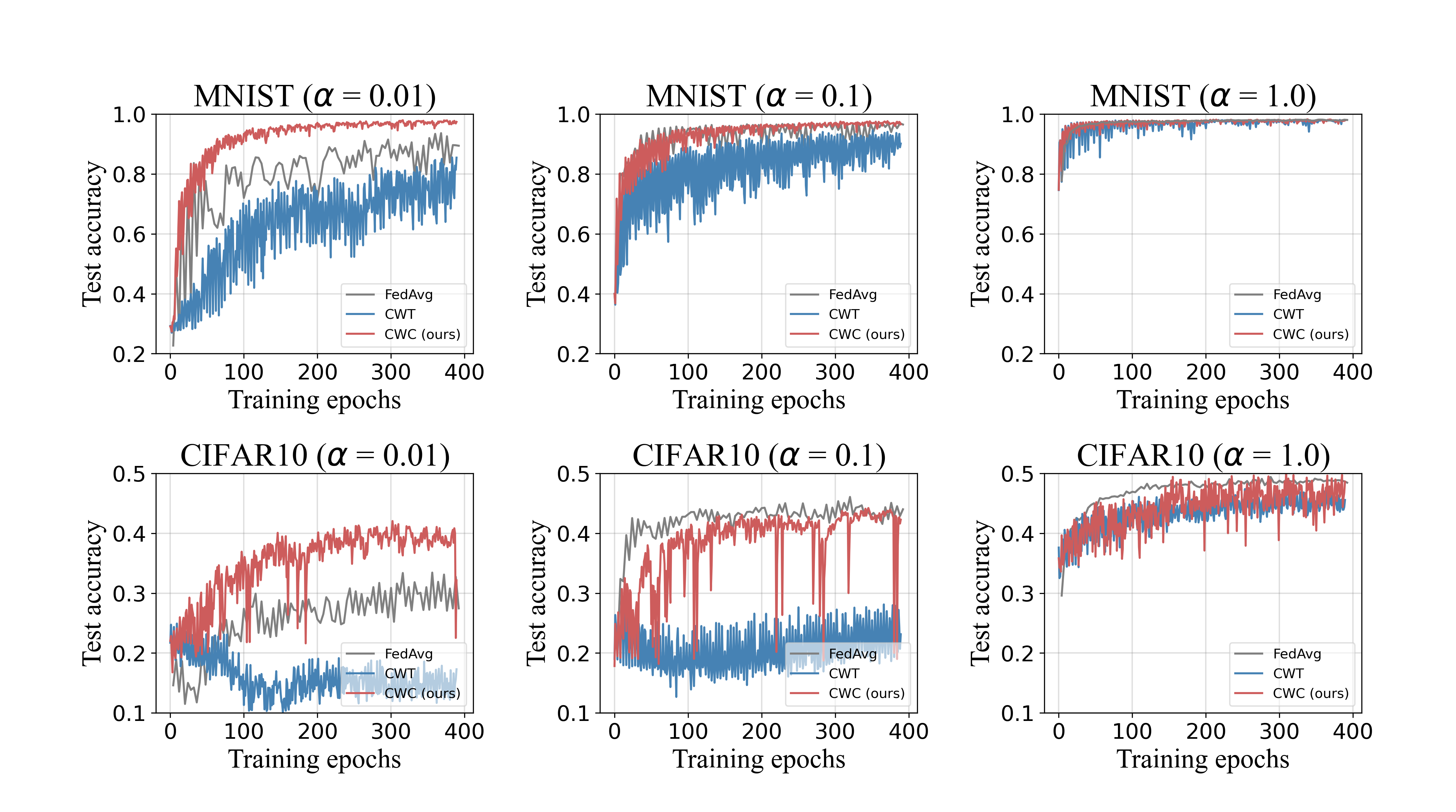}
    \caption{
    Visualize the overall performance of three algorithms on MNIST and CIFAR10 under three non-IID settings (i.e., $\alpha=0.01$, $\alpha=0.1$, $\alpha=1.0$ from left to right). The curves plot the classification accuracy on the balanced global test set alongside training epochs. It is important to note that serial federated learning is evaluated at the end of each training epoch, while parallel federated learning is evaluated every four training epochs (i.e., one communication round).
    }
    \label{fig:dirichlet}
\end{figure*}

In this section, we conduct a thorough evaluation of the effectiveness of CWC on three benchmarks: MNIST~\cite{mnist} and CIFAR10~\cite{cifar} with Dirichlet heterogeneous partitions, ISIC2018~\cite{isic,codella2019skin} skin disease classification also simulated with Dirichlet distribution, and a benchmark featuring extremely heterogeneous partitions. The results demonstrate that CWC consistently outperforms CWT significantly and achieves comparable or superior performance to FedAvg. Finally, we undertake an analytical study to investigate the influence of local training epochs and the consolidation factor on our CWC algorithm.

\subsection{Dirichlet Benchmark on MNIST and CIFAR10}

\begin{figure*}[t]
\centering
    \includegraphics[width=\textwidth]{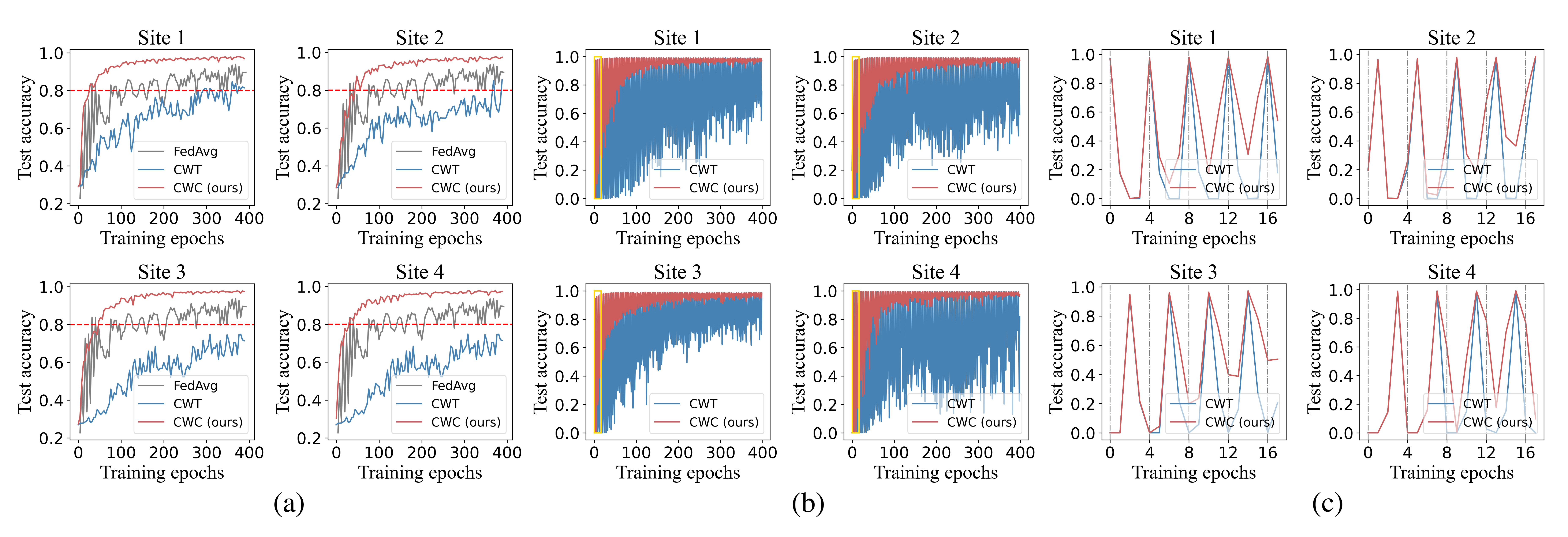}
    \caption{
    On the left, the plots display the classification accuracy on the balanced global test set based on the model that has just been updated on each site. In the middle, the plots illustrate the classification accuracy on the site-specific private test set across the entire training trajectory, while on the right, an amplification of the middle's initial stage is presented.
    }
    \label{fig:detials}
    \vspace{-4mm}
\end{figure*}

\textbf{Implementation Details.} We initiate our assessment by evaluating the effectiveness of our algorithm in image classification tasks using the MNIST and CIFAR10 datasets. To introduce data heterogeneity, we incorporate a Dirichlet distribution~\cite{dirichlet}, utilizing the concentration parameter $\alpha$ to control the level of heterogeneity. A smaller $\alpha$ value indicates a more imbalanced distribution of data across clients. In our experiments, we explore $\alpha$ values from the set $\{0.01, 0.1, 1.0\}$. For this study, we generate a total of four simulated clients. To ensure that the local test set and local train set within the same client are derived from the same distribution, we initially merge the train and test sets of MNIST and CIFAR10. Subsequently, we simulate four heterogeneous data silos for each client based on the specified concentration parameter $\alpha$. Within these silos, we further balance the data into local train and local test sets. To assess the overall model performance, we aggregate all local test sets into a global test set. This global test set maintains a balanced class distribution, enabling an unbiased evaluation of the model's performance. We employ a compact multi-layer perceptron (MLP) with two hidden layers, featuring 256 units for MNIST and 1024 units for CIFAR10. Each layer incorporates ReLU nonlinearities, and the model utilizes a standard categorical cross-entropy loss function. For MNIST, we set $\sigma=0.1$ and $\gamma=0.5$, while for CIFAR10, we set $\sigma=10.0$ and $\gamma=0.9$. And we employ SI~\cite{SI} as the important weight estimator. To ensure a fair comparison between serial FL (i.e., CWT, CWC) and FedAvg, we standardize the evaluation in terms of communication rounds, as each client receives and transmits the updated model weights once in both serial and parallel FL during each communication round, which suggests same communication cost. However, given that serial FL trains the global model sequentially, we conduct a total of $K=4$ evaluations, in contrast to the single evaluation performed with the aggregated model in parallel FedAvg. For optimal convergence, we optimize our network one epoch per communication round and for a total of 100 rounds (equivalent to 400 training epochs). To achieve robust absolute performance with a reduced number of epochs, we utilize the adaptive optimizer Adam~\cite{adam}.

\textbf{Main Results.} Fig.~\ref{fig:dirichlet} and Table~\ref{tab:main} illustrates a comparative analysis of overall performance among FedAvg, CWT, and Our CWC. To begin, we examine CWC against CWT, the serial approach without consolidation. Notably, CWC consistently outperforms CWT in all the three non-IID scenarios, alleviating performance fluctuations faced by CWT and achieving higher convergence scores. As data heterogeneity increases (from $\alpha=1.0$ to $\alpha=0.01$), CWT encounters challenges, particularly in addressing the exacerbated issue of catastrophic forgetting due to more heterogeneous distribution. The oscillation amplitude increases, resulting in lower convergence values. Moreover, in CIFAR10, CWT struggles to learn, evidenced by declining accuracy with increasing training epochs. In contrast, CWC, incorporating a cyclical consolidation mechanism, overcomes the inefficiency in the iterative process of learning and forgetting. This is evident in its more steadily growing trend and eventual convergence to higher accuracy. The figure also demonstrates the robustness of CWC to increasing data heterogeneity. As heterogeneity increases, the performance loss is moderate, in contrast to CWT and FedAvg. Remarkably, with cyclical weight consolidation, the serial federated learning approach becomes comparable to FedAvg and even outperforms it in highly heterogeneous scenarios ($\alpha=0.01$).

\textbf{In-depth Analysis.} The left figure of Fig.~\ref{fig:detials} displays the learning dynamics of the just-updated model for each site. As FedAvg does not have the site-specific evaluation, we substitute it with the global version. It can be observed that the CWT curve exhibits less oscillation compared to the plot showing its entire training trajectory in Fig.~\ref{fig:dirichlet}, reinforcing the idea that the fluctuation behavior in serial federated learning stems from the transition between sites. In contrast, CWC remains fairly stable, as illustrated by a similar trend in all four sites. This observation validates that a model trained with cyclical weight consolidation preserves the previously acquired knowledge when fitting the data from another dissimilar distribution, confirming the efficacy of our consolidation mechanism. The middle figure of Fig.~\ref{fig:detials} illustrates the classification accuracy on the site-specific private test set along the entire training trajectory, while the right figure of Fig.~\ref{fig:detials} shows the amplification of the middle's initial stage. As observed, initially, for both serial federated learning algorithms, the on-site test score experiences an extreme fluctuation trend: after fitting the on-site training samples, it reaches up to 100\%; however, when the model transitions and adapts to the data of the next site, the accuracy suddenly drops to a very low level. As the model undergoes more training epochs, the lowest value it drops to in a communication round exhibits a growing trend, indicating its ability to preserve a certain amount of the previously learned knowledge as it continually adapts to other distinct distributions. However, in CWT, the learning efficiency is extremely low, taking hundreds of communication rounds to converge to the level of 50\%. Moreover, the marginal gain also decreases over time. In contrast, CWC showcases its superiority with a faster convergence speed, proving its capacity to preserve previously acquired knowledge and its plasticity to learn knowledge from other data distributions.

\subsection{ISIC2018 Benchmark}
\begin{table}
\caption{Classification accuracy on the global test set under three heterogeneous settings. The final results are determined by selecting the best scores within 100 communication rounds. The results for MNIST are presented above, while those for CIFAR10 are displayed below.}
\centering
\scalebox{0.85}{

\begin{tabular}{l|c|c|c|c}
\toprule

  & \textbf{$\alpha=0.01$} & \textbf{$\alpha=0.1$} & \textbf{$\alpha=1.0$} & Mean \\
\midrule
FedAvg & 93.64 & 97.10 & 98.16 & 96.30 \\
CWT & 85.54 & 94.58 & 98.15 & 92.76 \\
CWC (ours) & \textbf{97.97} & \textbf{97.66} & \textbf{98.26} & \textbf{97.96} \\
\midrule
FedAvg & 33.36 & \textbf{46.07} & 49.40 & 42.94 \\
CWT & 24.92 & 28.38 & 47.38 & 33.56 \\
CWC (ours) & \textbf{42.04} & 44.22 & \textbf{50.50} & \textbf{45.59} \\

\bottomrule
\end{tabular}}

\label{tab:main}
\vspace{-1mm}
\end{table}
\begin{table}[t]
\caption{Balanced classification accuracy on the global test set under three heterogeneous settings. The final results are determined by selecting the best scores within 150 communication rounds.}
\centering
\scalebox{0.85}{

\begin{tabular}{l|c|c|c|c}
\toprule

  & \textbf{$\alpha=0.1$} & \textbf{$\alpha=1.0$} & \textbf{$\alpha=10.0$} & Mean \\
\midrule
FedAvg & 39.28 & 43.05 & 46.37 & 42.90 \\
CWT & 34.70 & 48.47 & 49.58 & 44.25 \\
CWC (ours) & \textbf{45.67} & \textbf{56.39} & \textbf{54.75} & \textbf{52.27} \\
\bottomrule
\end{tabular}}

\label{tab:isic}
\vspace{-2mm}
\end{table}
\begin{figure}[t]
\centering
    \includegraphics[width=0.7\linewidth]{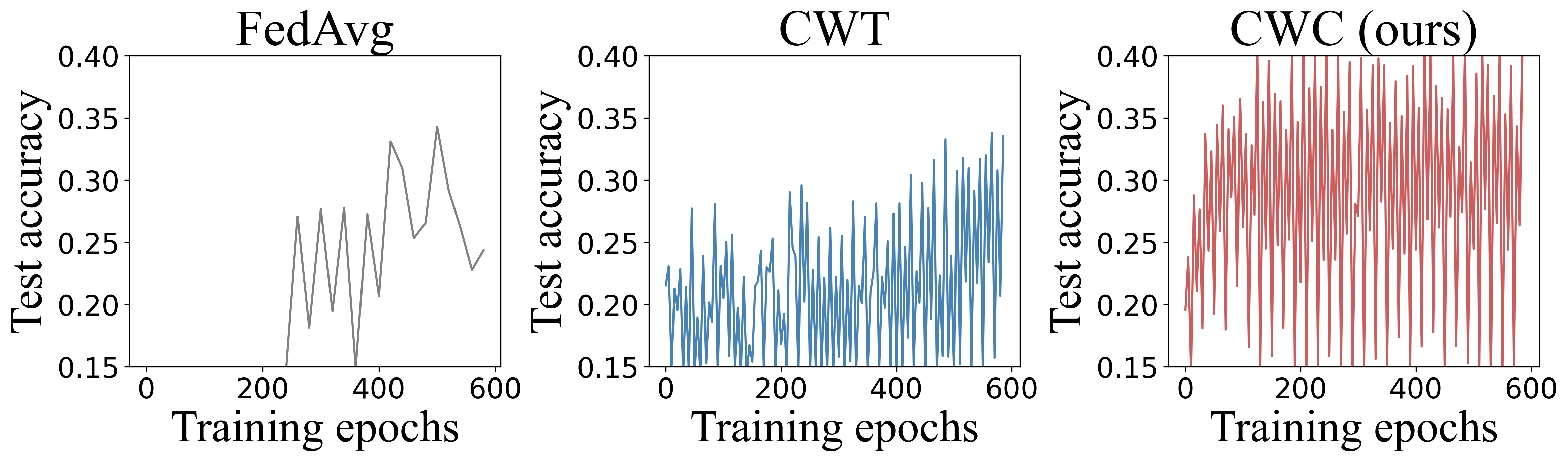}
    \caption{
    Balanced accuracy on the global test set along with training epochs under the setting of $\alpha=0.1$. From left to right, we depict the learning dynamics of FedAvg, CWT, and our CWC.
    }
    \label{fig:isic}
    \vspace{-2mm}
\end{figure}

To assess whether cyclical weight consolidation can also mitigate catastrophic forgetting in more complex datasets and larger models, we conducted experiments on a skin disease classification task based on ISIC2018~\cite{codella2019skin,isic}, containing over 10,000 images covering 7 disease categories. Specifically, we trained a ResNet50~\cite{resnet} without pretraining. Similar to previous benchmarks, we employed the Dirichlet distribution with a total of four clients to simulate various non-IID scenarios. However, in this benchmark, we used $\alpha \in \{0.1, 1.0, 10.0\}$. Given the imbalanced classes in ISIC2018, following~\cite{codella2019skin}, we adopted balanced accuracy (BACC) as our evaluation metric. And we set $\sigma=10.0$ and $\gamma=0.95$. To ensure proper convergence, we trained one epoch per communication round and for 150 communication rounds in total. 

Table~\ref{tab:isic} showcases the superiority of CWC against CWT and FedAvg in this scenario. Interestingly, FedAvg achieves the lowest performance in this class-imbalanced classification task, with a score of 42.9\%, slightly lower than CWT's 44.25\%. Evidently, CWC surpasses both with a significant margin. Fig.~\ref{fig:isic} plots the learning dynamics under the setting of $\alpha=0.1$. It is observed that significant fluctuation is present in all three algorithms, as the noise introduced by the highly imbalanced class distribution is unavoidable.

\subsection{Benchmark on Extremely Heterogeneous Distribution}
 
\begin{figure}[t]
\centering
    \includegraphics[width=0.5\linewidth]{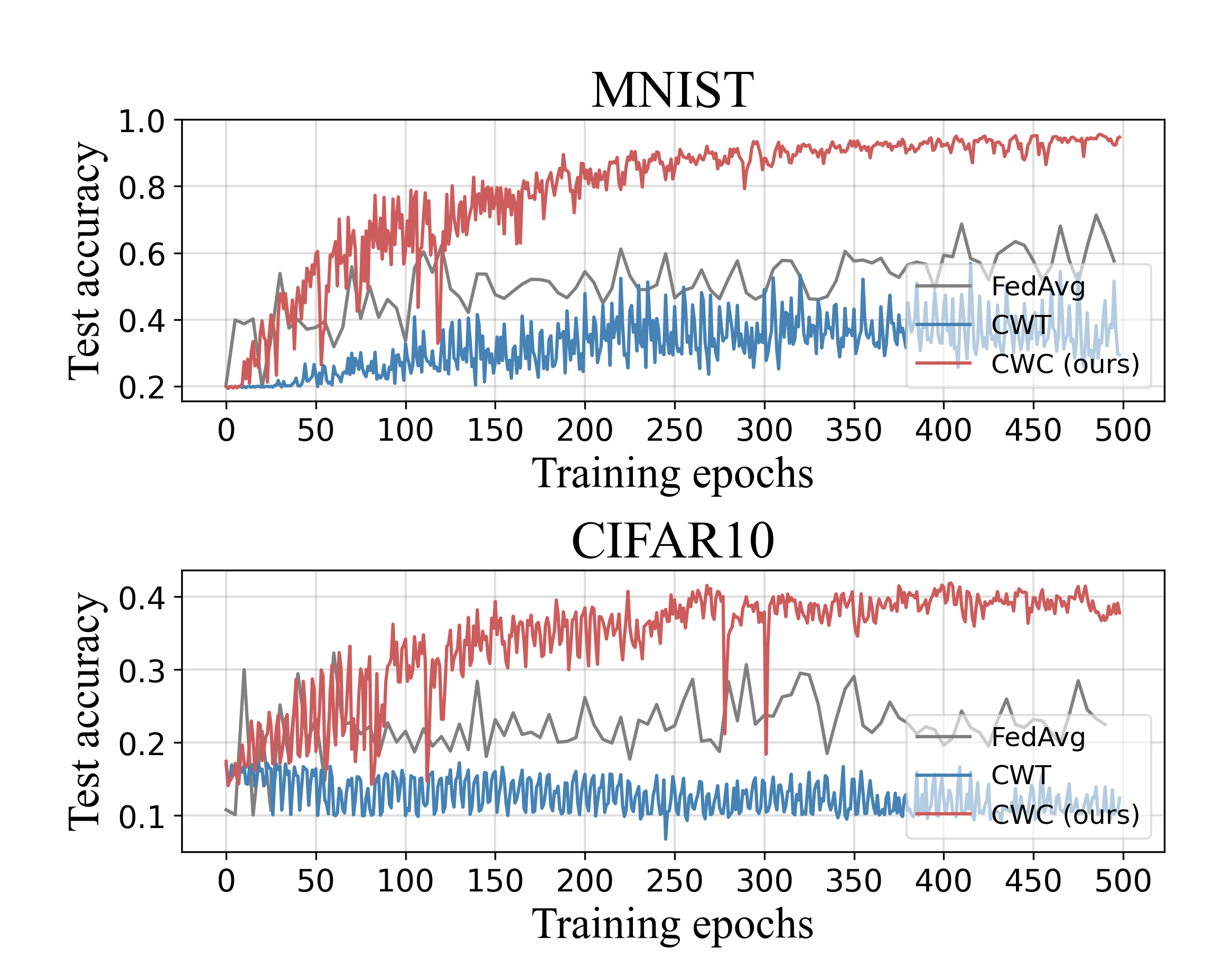}
    \caption{
  Test accuracy on the global test set along with training epochs for FedAvg, CWT, and our CWC. The upper part displays the results for MNIST, while the lower part shows the results for CIFAR10.
     }
    \label{fig:split}
 \vspace{-4mm}
\end{figure}

In this section, we evaluate CWC on distributions characterized by extreme heterogeneity. The complete MNIST and CIFAR10 training datasets are partitioned into five subsets of consecutive categories, with each client assigned two consecutive categories. And we utilized the official balanced test set for model performance evaluation. Unlike the Dirichlet benchmark, there is no class overlap in the samples from different sites, posing a significant challenge for FL algorithms. The remaining configurations are maintained consistent with the Dirichlet benchmark. Fig.~\ref{fig:split} illustrates the model performance on the official test set across different algorithms. CWC stands out by outperforming CWT and FedAvg by a considerable margin. Given that each site processes samples from mutually exclusive categories, different sites share minimal overlapping knowledge. Therefore local data optimizes the model in distinct directions, leading to substantial forgetting in CWT and notable client drift in FedAvg. Consequently, these two baseline algorithms struggle to generalize effectively. Contrastingly, CWC maintains strong performance in this benchmark, even comparable to the results obtained in Dirichlet experiments, albeit with a slight lag. The resilience of CWC under extreme heterogeneity highlights its ability to adapt and learn from diverse local data, overcoming challenges associated with the absence of shared knowledge across sites.

\subsection{Impact of Local Training Epochs}

\begin{figure}[t]
\centering
    \includegraphics[width=0.6\linewidth]{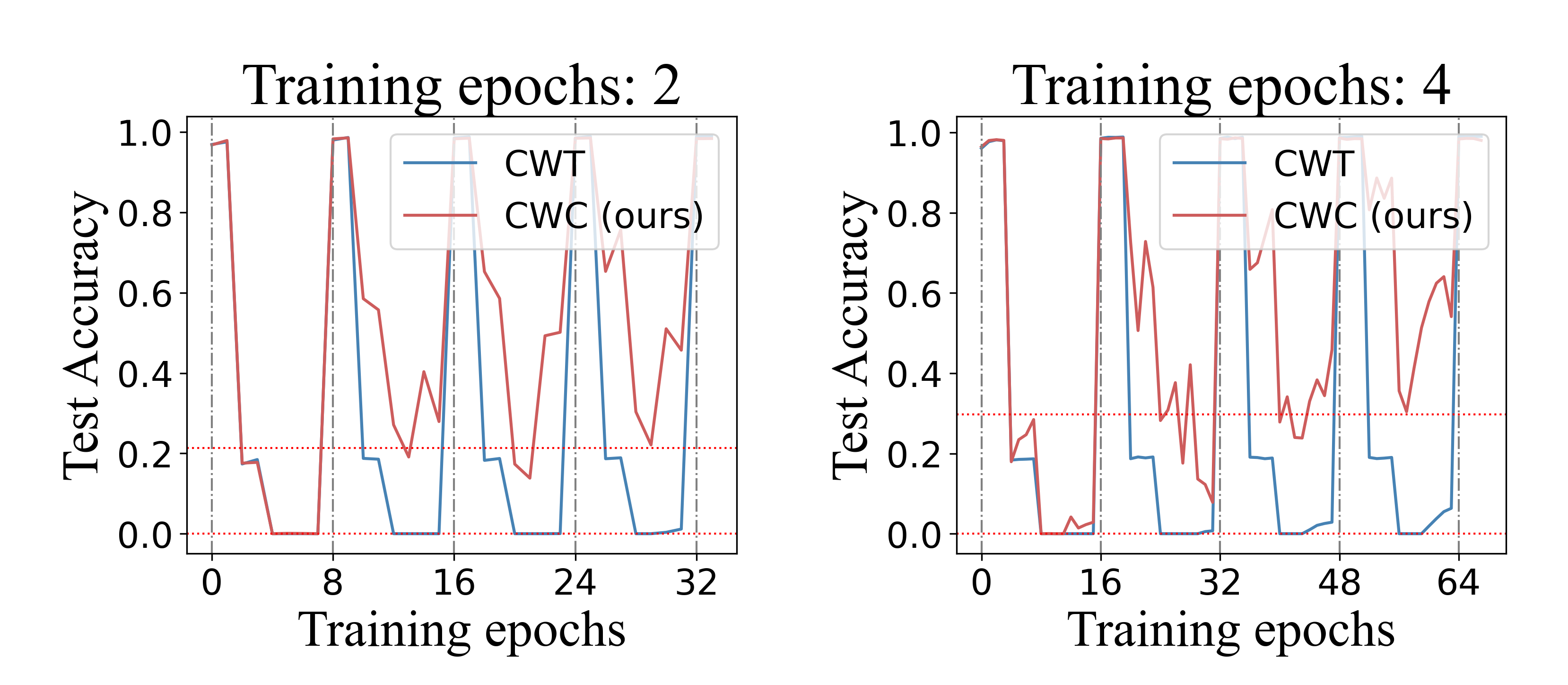}
    \caption{
  The learning dynamics in the startup stage under the setting of two local training epochs (left) and four local training epochs (right).
    }
    \label{fig:epochs}
    \vspace{-2mm}
\end{figure}

\begin{table}
\caption{Number of communication rounds (along with training epochs) needed to achieve 75\% accuracy on MNIST Dirichlet ($\alpha=0.01$) for varying local training epochs}
\centering
\scalebox{0.8}{
\begin{tabular}{l|cccc}
\toprule
\textbf{Local Epochs} & \textbf{1} & \textbf{2} & \textbf{4} & \textbf{8}\\
\midrule
CWT & 37 (148) & 41 (321)& 50 (787) & 39 (1240) \\
CWC (ours) & 6 (21) & 4 (25) & 3 (33) & 3 (66) \\
\bottomrule
\end{tabular}}

\label{tab:epochs}
\vspace{-1mm}
\end{table}
In parallel FL, the number of local training epochs is increased to reduce communication costs~\cite{fedavg}. The model can converge to the same performance with fewer communication rounds at the expense of more intense local computing. However, as pointed out in~\cite{sheller1, sheller2}, the benefit of increasing local training epochs in CWT is limited. CWT's performance is sensitive to the number of local training epochs because a longer exposure to heterogeneous data demands a stronger ability to preserve previously acquired knowledge. An increased number of local training epochs pose a significant challenge to its inherently mediocre memory. Table~\ref{tab:epochs} displays the number of communication rounds (along with training epochs) required to reach 75\% accuracy on MNIST Dirichlet ($\alpha=0.01$) for different local training epochs. Observing the results, an increase in the amount of local computation has no effect on the convergence speed in terms of CWT. In fact, it takes more communication rounds to reach the same preset accuracy (communication rounds increase to 41 and 50 under 2 and 8 epochs, respectively). In contrast, CWC experiences a boost in convergence speed with increased local epochs, similar to FedAvg. This is further demonstrated in Fig.~\ref{fig:epochs}, which shows the startup learning dynamics under the setup of 2 and 4 local epochs: After experiencing a total of four communication rounds, CWC's test accuracy converges to a higher level with a doubling in local training epochs.

\subsection{Impact of Consolidation Factor}

\begin{figure}[t]
\centering
    \includegraphics[width=0.6\linewidth]{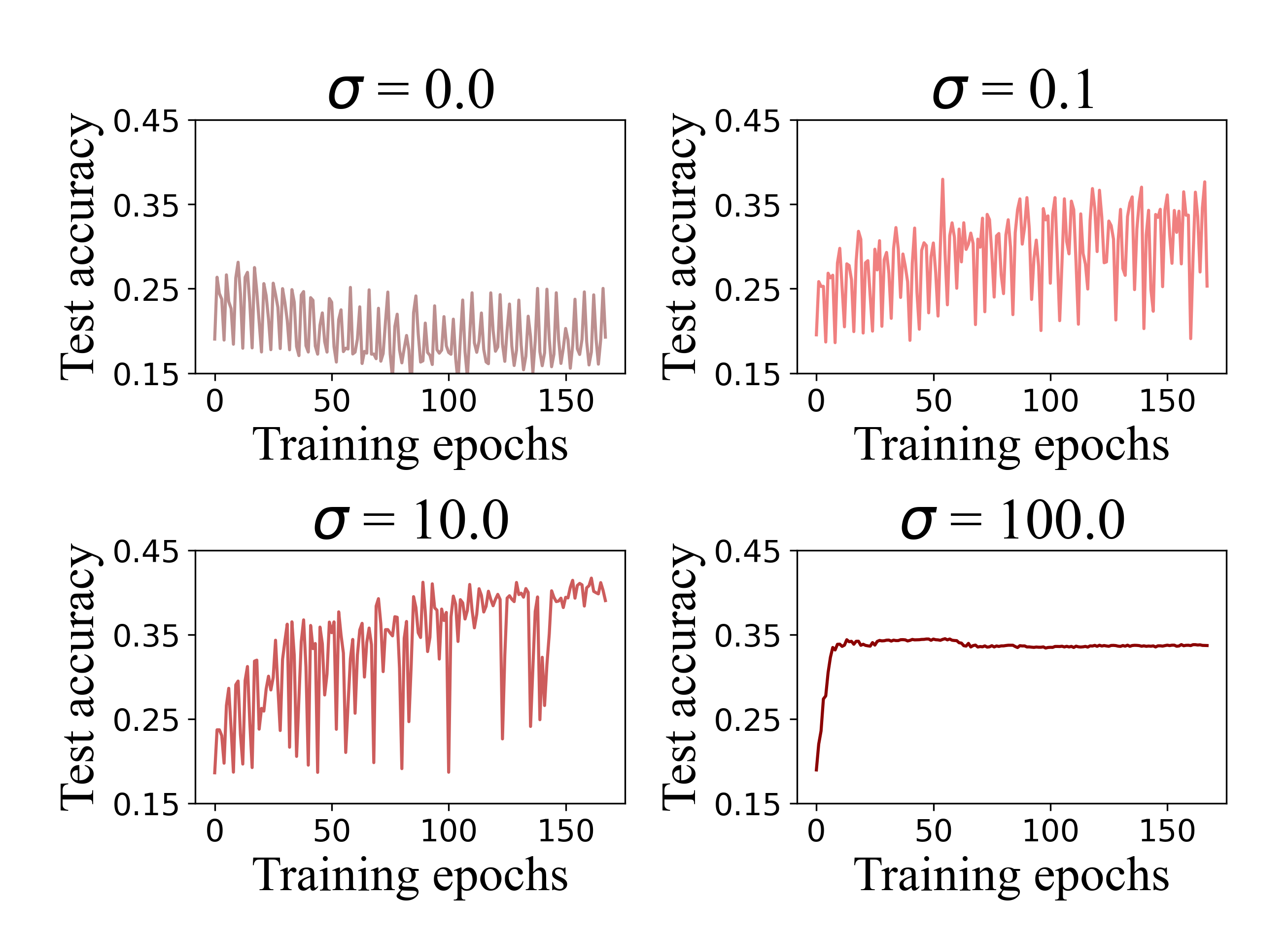}
    \caption{
    Test accuracy along with training epochs for CIFAR10 Dirichlet ($\alpha=0.1$) under four consolidation factors (i.e., 0.0, 1.0, 10.0, and 100.0 in the order from top to bottom, left to right).  
    }
    \label{fig:consolidation}
    \vspace{-4mm}
\end{figure}

In this investigation, we explore the impact of the consolidation factor $\sigma$ on CIFAR10 Dirichlet ($\alpha=0.1$). The test accuracy, plotted against training epochs, is depicted in Fig.~\ref{fig:consolidation} for four consolidation factors (i.e., 0.0, 1.0, 10.0, and 100.0). As expected, CWC without a consolidation mechanism fails to generalize to the global test set. With an increased consolidation effect (using $\sigma=1.0$), the model's performance exhibits an upward trend over time, accompanied by noticeable oscillation. When $\sigma$ is set to 10.0, the oscillation is attenuated, resulting in improved convergence performance. Further increasing $\sigma$ to 100.0 eliminates oscillation, but the curve converges to a lower level. From these observations, we can infer that the consolidation factor plays a crucial role in balancing resilience to previously acquired knowledge and plasticity to adapt to new knowledge. Achieving an optimal performance requires finding a delicate balance in between.
\section{Conclusion}

In this study, we propose cyclical weight consolidation to address catastrophic forgetting in serial FL. Our extensive evaluations demonstrate the effectiveness of CWC in mitigating the fluctuation behavior of the original serial FL approach and consistently achieving higher convergence performance. We illustrated that CWC can adapt to more complex datasets and larger models, as evidenced by its success in a skin disease classification task. Notably, CWC outperformed CWT and FedAvg in extremely heterogeneous distributions, showcasing its robustness in challenging scenarios where other methods fail. Furthermore, our results indicate that CWC achieves performance comparable to parallel FL, marking a significant milestone for serial FL. The capability of CWC to boost convergence by intensifying local computation provides an additional advantage not achievable in CWT. We conducted a thorough analysis of the impact of the consolidation factor, demonstrating that optimal performance in CWC relies on striking a delicate balance between resilience to prior knowledge and adaptability to incorporate new information. Overall, our findings underscore the potential of CWC as a valuable tool in serial FL, offering a robust and efficient solution to the challenges posed by non-IID data distributions and knowledge retention across diverse sites.

\bibliographystyle{unsrtnat}
\bibliography{main}

\end{document}